\documentclass[11pt]{article}

\usepackage{amsmath, amssymb}
\usepackage{graphicx}
\usepackage{booktabs}
\usepackage{longtable}
\usepackage{hyperref}
\usepackage{geometry}
\title{
FLOW: A Feedback-Driven Synthetic Longitudinal Dataset of Work and Wellbeing
}

\author{
Wafaa El Husseini \\
\href{https://orcid.org/0000-0003-2344-5769}{ORCID: 0000-0003-2344-5769} \\
\texttt{elhusseini.wafaa95@gmail.com}
}

\date{}

\begin{document}

\maketitle


\begin{abstract}
Access to longitudinal, individual-level data on work-life balance and wellbeing
is limited by privacy, ethical, and logistical constraints. This poses challenges
for reproducible research, methodological benchmarking, and education in domains
such as stress modeling, behavioral analysis, and machine learning.

We introduce \textbf{FLOW}, a synthetic longitudinal dataset designed to model
daily interactions between workload, lifestyle behaviors, and wellbeing.
FLOW is generated using a rule-based, feedback-driven simulation that produces
coherent temporal dynamics across variables such as stress, sleep, mood, physical
activity, and body weight. The dataset simulates 1{,}000 individuals over a
two-year period with daily resolution and is released as a publicly available
resource.

In addition to the static dataset, we describe a configurable data generation
tool that enables reproducible experimentation under adjustable behavioral and
contextual assumptions. FLOW is intended as a controlled experimental environment
rather than a proxy for observed human populations, supporting exploratory
analysis, methodological development, and benchmarking where real-world data
are inaccessible.
\end{abstract}

\section{Introduction}

Understanding how work demands, lifestyle behaviors, and wellbeing interact over
time is central to research in occupational health, behavioral science, and data
science. Foundational models in occupational stress research emphasize the role
of job demands, control, and reward structures in shaping strain and health
outcomes \cite{karasek1979job, siegrist1996adverse}. Capturing these dynamics at
fine temporal resolution is important because stress, sleep, physical activity,
and mood can influence one another through short-term feedback mechanisms.

Most existing datasets in this domain rely on surveys, administrative records, or
experience sampling methods. National time-use studies and related survey efforts
have provided valuable population-level views of time allocation and wellbeing,
but are often cross-sectional or limited in temporal granularity
\cite{gershuny2014time, hamermesh2005timing}. Experience sampling and ecological
momentary assessment can capture higher-frequency signals, but such data are
costly to collect, difficult to scale, and rarely released openly due to privacy
and ethical constraints \cite{shiffman2008ecological}. As a result, publicly
available longitudinal datasets suitable for open benchmarking and education
remain limited.

In parallel, data-driven approaches to modeling mental health and wellbeing have
increasingly leveraged smartphone sensing and digital traces
\cite{wang2014studentlife, saeb2015mobile}. While these sources can enable
predictive modeling, they are frequently proprietary or restricted and therefore
difficult to use for reproducible research and community benchmarking.

Synthetic data offers a complementary approach to addressing these limitations.
Recent work on tabular data synthesis has explored generative models such as
GAN-based approaches for producing statistically similar records
\cite{park2018data, xu2019modeling}. However, many synthetic tabular methods
prioritize distributional similarity over interpretable temporal structure,
which is particularly important in longitudinal human-centered domains.
Simulation-based and agent-based modeling approaches provide an alternative that
can encode mechanism-driven dynamics and temporal coherence
\cite{bonabeau2002agent, auchincloss2012agent}, but are less commonly released as
reusable datasets compatible with standard analysis workflows.

In this paper, we introduce \textbf{FLOW}, a synthetic longitudinal dataset
designed to model daily work-life balance and wellbeing dynamics through
interpretable, rule-based mechanisms. FLOW emphasizes feedback-driven temporal
structure, where daily outcomes influence subsequent states rather than being
independently sampled. Professional context and interventions are incorporated
as modifiers of daily conditions rather than as causal policy drivers.

The contributions of this work are threefold:
\begin{itemize}
    \item We present a publicly available synthetic dataset with daily resolution,
    simulating 1{,}000 individuals over two years.
    \item We document a transparent, rule-based generation process that encodes
    feedback loops between workload, lifestyle behaviors, and wellbeing.
    \item We outline a configurable generator tool that supports reproducible
    experimentation under adjustable modeling assumptions.
\end{itemize}

FLOW is intended as a methodological and exploratory resource rather than a
substitute for real-world occupational health data. By making assumptions
explicit and controllable, the dataset aims to support responsible, reproducible
research and education in settings where access to empirical longitudinal data
is limited.

\section{Related Work}

Research on work-life balance, stress, and wellbeing has traditionally relied on
survey-based and observational data sources. Large-scale surveys such as national
time-use studies \cite{gershuny2014time, hamermesh2005timing} and occupational
health questionnaires \cite{karasek1979job, siegrist1996adverse} have provided
important population-level insights into working conditions, stress, and health
outcomes. However, these datasets are typically cross-sectional or sparsely
sampled over time, limiting their suitability for fine-grained longitudinal
analysis.

Experience sampling and ecological momentary assessment methods have been used
to capture short-term fluctuations in mood and stress \cite{shiffman2008ecological},
but such data are costly to collect, difficult to scale, and rarely released in
a form suitable for open research due to privacy concerns. As a result, publicly
available datasets that combine daily temporal resolution, long observation
periods, and individual-level behavioral variables remain scarce.

In parallel, machine learning research on mental health and wellbeing has
increasingly relied on proprietary data sources, such as workplace monitoring
tools or digital health platforms \cite{wang2014studentlife, saeb2015mobile}.
While these datasets enable predictive modeling, they are typically inaccessible
to the broader research community and cannot be redistributed for benchmarking
or educational use.

Synthetic data generation has emerged as a complementary approach to address
privacy and access limitations. Prior work has focused largely on tabular data
synthesis for anonymization and data sharing, using techniques such as Bayesian
networks, generative adversarial networks, or variational autoencoders
\cite{xu2019modeling, park2018data}. These approaches emphasize marginal
distribution matching but often lack explicit temporal structure or interpretable
behavioral mechanisms, particularly for longitudinal human-centered data.

Simulation-based and agent-based models have been used to study human behavior
and health dynamics \cite{bonabeau2002agent, auchincloss2012agent}, offering
greater interpretability and temporal coherence. However, such models are rarely
released as reusable datasets compatible with standard data analysis workflows,
and are often domain-specific or tightly coupled to particular research
questions.

FLOW builds on these strands of work by providing a publicly available synthetic
longitudinal dataset that encodes interpretable feedback loops between workload,
stress, lifestyle behaviors, and wellbeing. Rather than aiming to replicate any
specific real-world population, FLOW is designed as a controlled experimental
environment that supports exploratory analysis, methodological development, and
reproducible benchmarking under transparent assumptions.

\section{Design Philosophy}

FLOW is designed as a synthetic, rule-based simulation of daily human behavior, with the goal of supporting exploratory analysis, methodological
experimentation, and benchmarking in the context of work-life balance and
wellbeing. Rather than attempting to replicate any specific real-world survey
or population, the dataset encodes a set of interpretable assumptions about how
workload, lifestyle habits, and stress interact over time.

\subsection{Rule-Based Rather Than Random Generation}
Central to the design of FLOW is the presence of feedback loops, where daily outcomes influence subsequent states rather than being independently sampled.
Unlike purely stochastic synthetic datasets, the data generation process follows
explicit behavioral rules and feedback mechanisms. Variables such as stress,
sleep duration, mood, and energy are not sampled independently; instead, they
are generated through conditional relationships that reflect commonly observed
patterns in daily life. For example, increased workload raises stress levels,
which in turn may reduce sleep duration and negatively affect mood and focus.
This approach prioritizes internal coherence and temporal consistency over
exact statistical matching to external datasets.

\subsection{Longitudinal Focus and Temporal Dynamics}

Each individual in the dataset is simulated over an extended time horizon with
daily resolution. This longitudinal structure allows for the emergence of
short-term fluctuations as well as longer-term trends, such as cumulative sleep
debt or gradual changes in body weight. Weekly aggregation is included to
support multi-scale analysis while preserving the underlying daily dynamics.

\subsection{Professions as Contextual Modifiers}

Professional roles (e.g., managers, nurses, teachers) are modeled as contextual
modifiers rather than causal drivers. They influence baseline characteristics
such as typical workload, work schedules, or meeting frequency, but they do not
encode profession-specific policy changes or institutional interventions over
time. As a result, profession-level trends should be interpreted as contextual
differences in daily conditions rather than as evidence of structural or policy
effects.

\subsection{Interventions as Stochastic Exposures}

Wellbeing-related interventions, including vacation, sick leave, workload caps,
or lifestyle programs, are modeled as stochastic exposure events. Their presence
does not guarantee immediate or uniform improvements in stress or mood. Instead,
interventions interact with ongoing behavioral dynamics and may be attenuated
or amplified by competing factors such as workload cycles, seasonality, or
individual habits. This design choice reflects the uncertainty and variability
often observed in real-world behavioral responses.

\subsection{Stability and Variability of Variables}

Different variables are intentionally modeled with different degrees of
volatility. While stress, sleep duration, and daily activity may fluctuate
substantially over time, other attributes such as sleep quality or baseline mood
tend to evolve more slowly. This distinction is intended to prevent unrealistic
over-responsiveness and to preserve plausible temporal structure in the data.

\subsection{Trade-offs and Design Scope}

The dataset prioritizes interpretability, reproducibility, and ethical safety
over demographic representativeness or clinical validity. It is not calibrated
against real occupational health datasets and should not be used to infer causal
effects of employer policies or profession-specific interventions. Instead, it
serves as a controlled environment for exploring analytical methods, behavioral
hypotheses, and modeling approaches under transparent and adjustable assumptions.

\section{Data Generation Process}

The FLOW dataset is generated through a rule-based simulation pipeline that
models daily human behavior over time. The generation process is fully
deterministic given a random seed, ensuring reproducibility while allowing
controlled variability across individuals and scenarios. Rather than sampling
observations independently, the process simulates interacting variables through
explicit temporal and behavioral dependencies.

\subsection{Population Initialization}

The synthetic population is initialized by sampling a set of individual profiles,
each defined by static attributes such as age, profession, work mode (e.g.,
remote, onsite, hybrid), chronotype, and baseline lifestyle tendencies. These
attributes remain fixed throughout the simulation and serve as contextual
factors that influence daily behavior.

Individuals are assigned to broad professional categories that determine
baseline workload characteristics, such as typical working hours, meeting
frequency, or schedule regularity. These assignments affect initial conditions
but do not encode profession-specific policies or long-term structural changes.

\subsection{Daily Simulation Loop}

For each individual, the simulation proceeds on a daily time step across the
specified date range. On each day, variables related to work, lifestyle, and
wellbeing are updated sequentially based on the individual’s prior state, static
attributes, and exogenous factors.

Work-related variables, including working hours, meetings, and email volume,
are generated first and depend on both professional context and whether the day
is a workday. These variables influence perceived workload and contribute to the
individual’s stress level for that day.

Lifestyle variables such as physical activity, outdoor time, caffeine intake,
and diet quality are then generated as conditional responses to workload,
day-of-week effects, and seasonal patterns. For example, high workload or
critical pressure states may reduce exercise duration or diet quality, while
non-workdays allow for partial recovery.

\subsection{Stress, Sleep, and Mood Dynamics}

Stress acts as a central latent driver within FLOW and mediates interactions
between work and wellbeing. Daily stress levels are influenced by workload,
seasonality, and short-term workload cycles, and in turn affect sleep duration,
sleep quality, mood, energy, and focus.

Sleep duration is modeled as a function of stress, chronotype, caffeine intake,
and workday status. Sleep quality evolves more slowly and reflects both recent
sleep patterns and accumulated stress. Mood, energy, and focus are derived from
sleep, stress, and lifestyle factors, creating feedback loops in which daily
outcomes influence subsequent states.

\subsection{Weight and Energy Balance Modeling}

Body weight is modeled as a slowly evolving variable driven by daily energy
balance. For each individual, a basal metabolic rate is estimated using standard
physiological formulas, adjusted by an activity factor derived from physical
activity and work mode. Daily caloric intake is generated based on diet quality
and stress, with elevated stress increasing the likelihood of overeating.

Changes in body weight accumulate gradually over time, reflecting sustained
behavioral patterns rather than short-term fluctuations. This design enables the
emergence of long-term trends such as gradual weight gain under chronic stress.

\subsection{Temporal Aggregation}

In addition to daily records, FLOW provides weekly summary statistics computed
from daily observations. These include aggregated measures of stress, sleep,
activity, and wellbeing, as well as derived indicators such as sleep debt and job
satisfaction. Weekly aggregation supports multi-scale analysis while preserving
the underlying daily dynamics.

\subsection{Interventions and External Events}

Wellbeing interventions, including vacation, sick leave, workload caps, or
lifestyle programs, are introduced as external events with defined durations.
Interventions modify relevant daily variables probabilistically rather than
deterministically, reflecting heterogeneous individual responses.

Importantly, interventions do not override the existing behavioral dynamics.
Their effects may be attenuated or amplified by concurrent stressors, workload
cycles, or seasonal factors, and therefore do not guarantee immediate or uniform
improvements in wellbeing indicators.

\subsection{Reproducibility and Parameterization}

The entire generation process is controlled by a random seed, allowing exact
replication of datasets. Core parameters governing population size, time span,
stress sensitivity, intervention strength, and behavioral responsiveness can be
adjusted to generate alternative datasets under different assumptions. This
parameterization supports controlled experimentation and scenario analysis while
maintaining consistent structural logic.

\section{Dataset Description}

The FLOW dataset consists of multiple tabular files that together capture daily
wellbeing dynamics, contextual user attributes, temporal aggregates, and external
events. The dataset is released in comma-separated value (CSV) format and is
designed to support both modular and end-to-end analysis workflows.

\subsection{Scope and Scale}

FLOW simulates a population of 1{,}000 synthetic individuals over a two-year
period, from January 2024 to December 2025, with daily resolution. This includes
the 2024 leap year, resulting in 731 days per individual and a total of
731{,}000 daily observations. All records are fully synthetic and do not
correspond to real individuals.
The dataset scale enables both population-level analysis and fine-grained
longitudinal modeling at the individual level.
Across the dataset, \texttt{daily\_logs.csv} and \texttt{daily\_all.csv} each
contain 731{,}000 rows, corresponding to one observation per individual per
day. The remaining tables provide static or aggregated information and contain
substantially fewer rows.

\subsection{Core Tables}

\paragraph{Users.}
The \texttt{users.csv} table contains static demographic and contextual attributes
for each individual. These include age, sex, height, profession, work mode,
chronotype, baseline body mass index, and lifestyle tendencies. Each individual
is identified by a unique \texttt{user\_id} that is shared across all tables.

\paragraph{Daily Logs.}
The \texttt{daily\_logs.csv} table forms the core of the dataset and records daily
observations for each individual. Variables cover work-related factors (e.g.,
working hours, meetings), lifestyle behaviors (e.g., physical activity, caffeine
intake, diet quality), and wellbeing indicators (e.g., stress level, mood, sleep,
energy, focus, and body weight). Each row represents a single individual-day
combination.

\paragraph{Weekly Summaries.}
The \texttt{weekly\_summaries.csv} table provides weekly aggregations derived from
daily logs. It includes indicators such as average stress, sleep debt, job
satisfaction, anxiety and depression scores, average body weight, and counts of
low-quality diet days. Weekly summaries support medium-term trend analysis and
reduce noise in daily observations.

\paragraph{Interventions.}
The \texttt{interventions.csv} table records external wellbeing-related events,
including vacation, sick leave, workload caps, and lifestyle programs. Each
intervention is defined by a start date, end date, type, and intensity. These
events are intended to contextualize changes in daily behavior rather than
represent deterministic treatments.

\subsection{Denormalized Representation}

In addition to the normalized tables, FLOW includes a denormalized dataset,
\texttt{daily\_all.csv}, which contains one row per individual per day
(\(n = 731{,}000\) rows). This table combines static user attributes, daily
observations, weekly summary indicators, and flags for active interventions
into a single wide-format representation.

The denormalized table is provided to simplify exploratory data analysis and
machine learning workflows, where repeated joins across multiple tables may
introduce unnecessary complexity. While this format results in repeated
contextual information across rows, it preserves the original temporal
resolution and does not introduce additional synthetic signals.

\subsection{Variable Groups}

Across all tables, variables can be broadly grouped into the following categories:
\begin{itemize}
    \item \textbf{Contextual attributes:} demographic information, profession, and work mode.
    \item \textbf{Workload indicators:} working hours, meetings, emails, and perceived pressure.
    \item \textbf{Lifestyle behaviors:} physical activity, diet quality, caffeine intake, and screen time.
    \item \textbf{Wellbeing indicators:} stress, sleep, mood, energy, focus, and body weight.
    \item \textbf{Temporal aggregates:} weekly summaries and derived indicators.
    \item \textbf{External events:} wellbeing interventions and their durations.
\end{itemize}

\subsection{Data Format and Access}

All files are released in UTF-8 encoded CSV format and can be loaded using standard
data analysis tools. The dataset is publicly available on Kaggle and distributed
under an open license, enabling reuse, redistribution, and modification for
research and educational purposes.

\section{Validation and Sanity Checks}

To assess whether FLOW behaves as intended, we perform a series of internal
validation and sanity checks. These checks do not aim to establish external
validity with respect to real-world populations, but rather to verify internal
consistency, plausible ranges, and expected directional relationships between
key variables.

\subsection{Distributional Properties}

Across all daily observations (\(n = 731{,}000\)), core variables such as sleep
duration, working hours, physical activity, caffeine intake, and screen time
exhibit stable distributions with plausible ranges and moderate variance.
For example, average sleep duration centers around seven hours per night, while
working hours and activity levels vary meaningfully across individuals and days.
No variables exhibit implausible values or degenerate distributions.

\subsection{Expected Behavioral Relationships}

FLOW encodes several expected relationships between variables that emerge
consistently in aggregate analysis. Higher workload is associated with elevated
stress levels, while increased stress is linked to reduced sleep duration and
lower mood scores. Conversely, higher physical activity and better diet quality
tend to co-occur with improved mood, energy, and focus.

These relationships are not enforced deterministically at the individual level,
but arise statistically across the population due to the feedback-driven design
of the generation process.

\subsection{Temporal Consistency}

Because FLOW is simulated longitudinally, variables evolve smoothly over time
rather than fluctuating independently across days. Slowly changing variables
such as body weight and sleep quality exhibit gradual trends, while more reactive
variables such as stress or daily workload respond to short-term conditions.
This temporal structure supports time-series analysis and prevents unrealistic
day-to-day volatility.

\subsection{Heterogeneity Across Individuals}

The dataset exhibits substantial heterogeneity across individuals, driven by
differences in static attributes such as profession, work mode, chronotype, and
lifestyle tendencies. While population-level trends are observable, individual
trajectories vary considerably, reflecting the stochastic nature of daily
behavior and the influence of competing factors.

\subsection{Interventions and Attenuated Effects}

Periods during which wellbeing interventions are active do not uniformly result
in immediate improvements in stress or mood. This behavior is consistent with
the design of FLOW, where interventions act as probabilistic modifiers rather
than deterministic treatments. Their effects may be attenuated by concurrent
stressors, workload cycles, or seasonal influences, resulting in heterogeneous
outcomes.

\subsection{Limitations of Validation}

These validation checks confirm that FLOW exhibits coherent internal structure
and expected directional relationships. However, they do not imply calibration
to real-world populations or suitability for estimating causal or policy-level
effects. All validation is therefore interpreted in the context of the dataset’s
intended use as a synthetic, exploratory environment.

\section{Generator Tool}\label{sec:generator}

In addition to the static dataset release, FLOW is accompanied by a configurable
data generation tool designed to enable reproducible experimentation and
scenario analysis. This paper documents the first public version of FLOW; the
generator tool is currently in preparation and will be made available shortly
following the initial release of this manuscript.

\subsection{Motivation}

A single synthetic dataset represents only one realization of a broader set of
assumptions about human behavior and work-life dynamics. The generator tool is
intended to expose these assumptions explicitly and allow users to explore how
changes in parameter settings affect observed patterns. This approach supports
methodological robustness, sensitivity analysis, and controlled experimentation
that would not be possible with a fixed dataset alone.

\subsection{Parameterization}

The generator tool is designed around a set of interpretable parameters that
control population characteristics, behavioral responsiveness, and external
conditions. Planned parameters include, but are not limited to:

\begin{itemize}
    \item Population size and simulation time span.
    \item Random seed for full reproducibility.
    \item Baseline workload and stress sensitivity.
    \item Responsiveness of sleep, mood, and energy to stress.
    \item Strength and duration of wellbeing interventions.
    \item Distribution of work modes and professional contexts.
\end{itemize}

These parameters allow users to generate alternative datasets that share the
same structural logic as FLOW while differing in scale or behavioral intensity.

\subsection{Extended and Future Parameters}

Beyond the initial parameter set, the generator is designed to be extensible.
Future versions may introduce additional controls, such as alternative
intervention models, modified feedback strengths, or custom behavioral rules.
This extensibility is intended to support comparative studies and facilitate
experimentation under varying modeling assumptions.

\subsection{Relationship to the Static Dataset}

The dataset released with this paper corresponds to a specific parameter
configuration of the generator and serves as a reference instance for analysis
and benchmarking. Users are encouraged to treat this release as one example
within a broader family of datasets that can be generated under controlled and
transparent assumptions.

\subsection{Availability}

The generator tool will be released as open-source software and made publicly
available alongside the dataset. Documentation and example configurations will
be provided to support reproducible use. Links to the tool repository will be
added in a future revision of this manuscript.

\section{Intended Use and Limitations}

The FLOW dataset is intended to support exploratory data analysis, methodological
experimentation, and the development or evaluation of data-driven models in the
context of work-life balance and wellbeing. Typical use cases include feature
engineering, correlation analysis, time-series modeling, and benchmarking of
machine learning approaches under controlled and reproducible conditions.

The dataset is not designed to serve as a substitute for real-world occupational
health data, nor should it be used to draw clinical, diagnostic, or policy-level
conclusions. In particular, professional roles within FLOW function as contextual
modifiers of daily conditions rather than as causal drivers of structural or
institutional change. Consequently, profession-specific trends or intervention
effects should not be interpreted as evidence of real-world employer policies or
outcomes.

Wellbeing interventions in the dataset are modeled as stochastic exposure events
whose effects may be attenuated or amplified by concurrent factors such as
workload cycles, seasonality, or individual habits. Their presence does not imply
guaranteed improvements in stress, sleep, or mood.

The dataset is fully synthetic and was not calibrated against any specific
population or occupational cohort. While the generation process encodes
plausible behavioral relationships, it does not aim to achieve demographic
representativeness or clinical validity. These design choices prioritize ethical
safety, transparency, and reproducibility over realism at the individual or
policy level.

Users are encouraged to treat FLOW as a controlled experimental environment
rather than a proxy for observed human behavior, and to clearly state these
limitations when reporting results based on the dataset.

\section{Availability}

The FLOW dataset is publicly available on Kaggle under an open license and can be
accessed at \href{https://www.kaggle.com/datasets/wafaaelhusseini/worklife-balance-synthetic-daily-wellness-dataset/}{[FLOW Dataset on Kaggle]}. The release includes the normalized
tables and the denormalized \texttt{daily\_all.csv} file described in this paper.

The dataset is distributed under the Creative Commons Attribution–ShareAlike 4.0
International (CC BY-SA 4.0) license, which permits reuse, redistribution, and
adaptation with appropriate attribution and requires derivative datasets to be
released under the same license.

For citation purposes, users are encouraged to cite this paper in addition to
the Kaggle dataset page. A BibTeX entry will be provided on the dataset page
following the public release of this manuscript.

The accompanying data generation tool described in Section~\ref{sec:generator} is currently in
preparation and will be released as open-source software in a subsequent update.
Links to the repository and documentation will be added in a future revision of
this paper.
\subsection*{Citation}

If you use the FLOW dataset in academic work, please cite this paper as well as
the corresponding Kaggle dataset page.

{\small
\begin{verbatim}
@article{elhusseini2025flow,
  title={FLOW: A Feedback-Driven Synthetic Longitudinal Dataset of Work and Wellbeing},
  author={El Husseini, Wafaa},
  journal={arXiv preprint arXiv:XXXX.XXXXX},
  year={2025}
}
\end{verbatim}
}

\section{Conclusion and Future Work}

This paper introduced FLOW, a synthetic longitudinal dataset designed to model
daily interactions between workload, lifestyle behaviors, and wellbeing. By
combining rule-based generation, interpretable feedback mechanisms, and daily
temporal resolution, FLOW provides a controlled experimental environment for
exploratory analysis, methodological development, and reproducible benchmarking
in domains where real-world data are scarce or inaccessible.

The dataset is intentionally scoped to prioritize internal coherence,
reproducibility, and ethical safety over demographic representativeness or policy
fidelity. As such, FLOW should be understood as a modeling and experimentation
resource rather than a proxy for observed human populations.

Future work will focus on the public release of the accompanying data generation
tool, enabling users to generate alternative datasets under adjustable
assumptions and controlled parameter settings. Planned extensions include
additional intervention models, expanded parameterization of behavioral
responses, and support for scenario-based experimentation.

Further research directions include systematic sensitivity analysis of model
parameters, exploration of alternative feedback structures, and potential
calibration against aggregate statistics from real-world surveys where
appropriate. By making both the dataset and its underlying assumptions explicit,
FLOW aims to support transparent, responsible, and reproducible research on
work-life balance and wellbeing dynamics.

\bibliographystyle{plain}
\bibliography{references}

\end{document}